\documentclass{article}

    \usepackage[preprint,nonatbib]{neurips_2021}

\usepackage[utf8]{inputenc} 
\usepackage[T1]{fontenc}    
\usepackage{hyperref}       
\usepackage{url}            
\usepackage{booktabs}       
\usepackage{amsfonts}       
\usepackage{nicefrac}       
\usepackage{microtype}      
\usepackage{xcolor}         
\usepackage{graphicx}
\usepackage{tablefootnote}

\usepackage{enumitem}

\usepackage{algorithm}
\usepackage[noend]{algpseudocode}
\usepackage{tabularx}

\usepackage{csvsimple}

\title{Personalizing Pre-trained Models}

\usepackage{authblk}

\author{\textbf{Mina Khan}\textsuperscript{1} \quad  \textbf{P Srivatsa}\textsuperscript{*} \quad \textbf{Advait Rane}\textsuperscript{2*} \quad  \textbf{Shriram Chenniappa}\textsuperscript{2+} \\  \textbf{Asadali Hazariwala}\textsuperscript{2+} \quad \textbf{Pattie Maes}}
\affil[]{\textsuperscript{1}MIT Media Lab, Cambridge, MA, USA. \textsuperscript{2}Birla Institute of Technology, Goa, India.}

\begin{document}
\maketitle

\begin{abstract}
Self-supervised or weakly supervised models trained on large-scale datasets have shown sample-efficient transfer to diverse datasets in few-shot settings. 
We consider how upstream pretrained models can be leveraged for downstream few-shot, multilabel, and continual learning tasks.
Our model \emph{CLIPPER} (CLIP PERsonalized)  uses image representations from CLIP, a large-scale image representation learning model trained using weak natural language supervision. 
We developed a technique, called \emph{Multi-label Weight Imprinting} (MWI), for multi-label, continual, and few-shot learning, and
CLIPPER uses MWI with image representations from CLIP. 
We evaluated CLIPPER on 10 single-label and 5 multi-label datasets. Our model shows robust and competitive performance, and we set new benchmarks for few-shot, multi-label, and continual learning.
Our lightweight technique is also compute-efficient and enables privacy-preserving applications as the data is not sent to the upstream model for fine-tuning.
Thus, we enable few-shot, multilabel, and continual learning in compute-efficient and privacy-preserving settings.
\end{abstract}

\section{Introduction}
\footnotetext{ {\texttt{minakhan01@gmail.com}.
\url{https://github.com/PAL-ML/CLIPPER}
\textsuperscript{*,+}\textit{Equal Contribution}
}}

Data-efficiency and generalization are key challenges in deep learning, and representation learning has been at the heart of deep learning \cite{bengio2012deep}.
Recently, self-supervised or weakly supervised models have been leveraged to learn from large-scale uncurated datasets and have shown sample-efficient transfer to labeled tasks \cite{chen_big_2020, radford2021learning, henaff2020data, he2020momentum, devlin_bert_2019, radford2019language}.
However, commonly used transfer techniques, e.g., fine-tuning or distillation, do not currently support few-shot, multilabel, and continual learning.

Few-shot learning (FSL) has made great strides in the area of sample-efficient learning \cite{wang2020generalizing}. However, FSL models are pretrained on large, domain-specific, and expensive-to-label datasets and have not leveraged pretrained models to avoid training on large and domain-specific labeled datasets. Also, FSL methods do not outperform pretrained models when domain shift is present \cite{chen2019closer, kornblith2019better}. 

We consider the problem of enabling \textit{few-shot, multilabel, and continual learning} for real-world downstream tasks,  
and investigate combining representation learning from pretrained self-supervised or weakly supervised models with few-shot, multilabel, and continual learning techniques.

Our model \textbf{CLIPPER} {(CLIP PERsonalized)} uses image representations, i.e., embeddings, from CLIP, a weakly-supervised image representation learning model, for FSL.
Inspired by Weight Imprinting \cite{qi_low-shot_2018}, an FSL method, we develop an approach called Multilabel Weight Imprinting for few-shot, multilabel, and continual learning.
CLIPPER combines image embeddings from CLIP along with Multilabel Weight Imprinting for continual and multilabel few-shot learning. 

We evaluated CLIPPER on 10 single-label and 5 multi-label datasets. CLIPPER shows robust and competitive performance, e.g., it outperforms the state-of-the-art FSL method on MiniImagenet. We set benchmarks for few-shot, continual, and multilabel learning on several different datasets. 

We make 3 key contributions. 
\begin{enumerate}[topsep=0pt,itemsep=-1ex,partopsep=1ex,parsep=1ex]
    \item A new methodology combining the flexibility of few-shot learning methods with the sample-efficiency and generalizability of transfer learning methods using self-supervised or weakly supervised pretrained models. Our method eliminates the need for data- and compute-intensive pretraining on large, domain-specific, and labeled datasets for FSL. 
    \item A modified FSL technique, which enables few-shot, continual, and multilabel learning. Combined with our first contribution, our second contribution enables few-shot, continual, and multilabel learning using self-supervised or weakly supervised pretrained models. 
    \item Evaluations and benchmarks for few-shot, continual, and multilabel learning on 15 multilabel and single-label datasets, showing the robust and competitive performance of our technique.
\end{enumerate}

\section{Related Work}

\subsection{Few-shot Learning (FSL)}
There are three types of approaches for FSL: model-, metric-, and optimization-based. Unlike previous work, we use a pretrained weakly supervised model for data- and compute-efficient training.

Our Multilabel Weight Imprinting technique lies in the category of metric-based approaches \cite{koch2015siamese, snell_prototypical_2017, sung_learning_2018, vinyals_matching_2017}. 
More specifically, we use a prototype-based metric-learning approach, as they assign trainable proxies to each category and enable faster convergence via element-category comparison, instead of element-wise comparisons. 
Our work extends a previous FSL technique, called weight imprinting \cite{qi_low-shot_2018}. We not only use a pretrained base model \cite{qi_low-shot_2018}, instead of training a base network from scratch, but we also extend weight imprinting to enable multilabel and continual learning.
Other metric-based methods are complementary to our approach and our model can be further extended, e.g., with MatchingNet-like attention \cite{vinyals_matching_2017} or with RelationNet-like relation learning \cite{sung_learning_2018}.

Model-based methods \cite{santoro2016one, munkhdalai_meta_2017} use especially designed models for rapid parameter updates, and optimization-based techniques \cite{ravi_optimization_2016, finn_model-agnostic_2017, nichol2018first} adjust the optimization method to meta-learn efficiently. Recent research indicates that learning a good embedding model can be more effective than sophisticated meta-learning algorithms \cite{tian2020rethinking} and efficient meta-learning may be predominantly due to the reuse of high-quality features \cite{raghu2019rapid}. Nonetheless, these techniques, though relatively training-intensive, are complementary to our work and may be used to improve both upstream and downstream models.

\subsection{Self-supervised Representation Learning}

Self-supervised and weakly supervised models have been used in natural language processing \cite{dai2015semi, radford2018improving, devlin_bert_2019} and computer vision \cite{henaff2020data, he2020momentum, chen_simple_2020, chen_big_2020, radford2021learning} to learn from large-scale unlabeled or weakly labeled datasets. Though pre-training is still imperfect \cite{ericsson2020well, mahajan2018exploring}, pretrained models trained on large-scale datasets have shown robust and sample-efficient transfer to diverse tasks \cite{he2020momentum, henaff2020data, chen_big_2020, radford2021learning}.

Transfer learning is closely related to few-shot learning, but FSL does not use a pretrained method. Instead, FSL is trained and evaluated using the same distribution and does not necessarily outperform transfer learning when domain shift is present \cite{chen2019closer}. Transfer learning, however, could benefit from the specialized FSL techniques \cite{kornblith2019better}. Also, to the best of our knowledge, unlike our work, transfer learning using pretrained models has not been combined with multi-label and continual learning.

Like previous work, we use self-supervised learning via data augmentation to boost FSL \cite{gidaris2019boosting, qi_low-shot_2018}. 

\subsection{Multilabel and Continual Learning}
Continual learning techniques \cite{mai2021online} include regularization-based methods (e.g., Elastic Weight Consolidation), memory-based methods (e.g., Incremental Classifier and Representation Learning \cite{rebuffi2017icarl}), and parameter isolation (like Continual Neural Dirichlet Process Mixture). Previously used continual techniques, however, did not use pretrained models for few-shot, multilabel, and continual learning. 

Common multilabel classification techniques include ML-kNN, Multi-label DecisionTree, etc \cite{devkar2017survey}. Multi-Label Image Classification has also been done using knowledge distillation from weakly supervised detection \cite{liu2018multi}. However, none of the existing methods combine multilabel, continual, and few-shot learning, especially using pre-trained models. Several multilabel and continual learning techniques, nonetheless, are complementary to our work and can be extensions of our work.

\section{Approach}

\subsection{Desiderata}
We outline 3 desiderata for real-world computer vision applications.
First, \textbf{few-shot learning} so that the applications can start well in data-scarce scenarios and can also be customized and personalized for different needs. 
Second, \textbf{continual learning} to incrementally learn new information and avoid catastrophic forgetting, e.g., replacing of older classes when new ones are added. 
Third, \textbf{multilabel learning} as the right label may not be just one label but a subset of all the given labels, including 0 to all labels.
The multi-label case is important for not only assigning multiple labels to a particular data point but also for assigning zero labels, in case we get data points that we currently do not have labels for, i.e., the continual learning case. Continual learning often considers the addition of data points along with their respective labels. However, we consider the more realistic continual case when a point may be added even before their label is added and thus, the model needs to assign no label.

\subsection{Decisions}

We made the following three design choices to enable few-shot, multilabel, and continual learning.

\textbf{Pretrained base model}: FSL models are typically pre-trained on large domain-specific training sets, which contain examples not in the support/test set. The models are then trained and tested on support and test sets, which have the same classes. 
Large-scale, domain-specific, and labeled datasets, however, may not always be available in real-world settings. Also, FSL models trained on domain-specific sets may not generalize well to domain shifts \cite{hu2021well}. Large-scale self-supervised or weakly supervised models, on the other hand, learn good representations and can be fine-tuned for data-efficient and diverse downstream tasks \cite{chen_big_2020, radford2021learning}. We use pretrained models trained on diverse datasets as base models for FSL, instead of training base models from scratch on domain-specific datasets. As a result, unlike FSL methods, we only train with a support test, which we call train set.

\textbf{Weight Imprinting (WI):} Weight imprinting \cite{qi_low-shot_2018} is a FSL learning method that learns a linear layer on top of the embeddings, where the columns of the linear layer weight matrix are prototype embeddings for each class. Many self-supervised or weakly supervised models have been shown to learn linearly-separable embeddings using linear probes \cite{radford2021learning} and thus, a linear layer can be added to pretrained embeddings to classify different classes. Compared to transfer learning, which learns a fixed number of classes, weight imprinting adds new classes as new columns of the linear layer weight matrix, making adding classes computationally and conceptually simpler and avoiding catastrophic forgetting. Thus, weight imprinting supports prototype-based few-shot and continual learning.

\textbf{Sigmoids, not Softmax:} The original weight imprinting model uses softmax and thus is compatible with single-label classification. We replace the softmax with sigmoid activations for each class in weight imprinting to enable multi-label learning. Sigmoids also support an output of 0 labels for continual learning, i.e., when the label for a given data point has not yet been added to the label set.

\subsection{Details}

We created a multilabel version of weight imprinting \cite{qi_low-shot_2018}, called \textbf{Multilabel Weight Imprinting (MWI)}. Our model has two parts. First, an embeddings extractor, $\phi : \mathbb{R}^N \rightarrow \mathbb{R}^D$, maps input image $x \in \mathbb{R}^N$ to a $D$-dimensional embedding vector $\phi(x)$, followed by an $L_2$ norm. Second, a sigmoid function, $f(\phi(x))$, maps the embedding using sigmoid activations for each category.

\[f_{i}(\phi(x)) = \frac{1}{1-exp(-w_i^T \phi(x))} \] where $w_i$ is the $i$-th column of the weight matrix normalized to unit length (with no bias term).

Each column of the weight imprinting matrix is a template of the corresponding category. The linear layer computes the inner product between the input embeddings $\phi(x)$ and each template embedding $w_i$. The result represents `close-by' templates in the embedding space using a threshold function.

\[\hat{y} = sgn (w^T \phi(x) - \vartheta)\]
where $sgn$ is the sign function and $\vartheta$ is the threshold.

\section{Implementation}

\begin{figure}
    \centering
    \includegraphics[width=\linewidth]{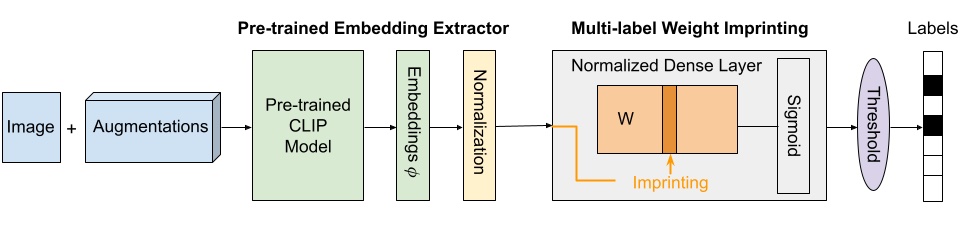}
    \caption{CLIPPER (CLIP-PERsonalized) uses embeddings from CLIP's vision transformer with Multilabel Weight Imprinting technique for few-shot, multilabel, and continual learning.}
    \label{fig:architecture}
\end{figure}

We share our implementation details below, model architecture in Fig \ref{fig:architecture}, and algorithm in Appendix.

\textbf{Embeddings Generator:} Weight imprinting \cite{qi_low-shot_2018} uses a base classifier trained on ``abundant'' labeled training samples. We replace the base classifier with CLIP (ViT B/32), a pretrained weakly supervised model \cite{radford2021learning}. We do not re-train or fine-tune the weights of the pretrained CLIP model. As shown in section 6 (Figure \ref{fig:fig1}), we compared embeddings from different supervised, self-supervised, and weakly supervised models, and chose CLIP because it had the best FSL performance using WI.

\textbf{Image Embeddings:} We embed images using CLIP's vision transformer and then use the normalized embeddings for multilabel weight imprinting. 
Compared to weight imprinting \cite{qi_low-shot_2018}, which used 64-dimensional embeddings, we use 512-dimensional embeddings from CLIP. Qi et al. \cite{qi_low-shot_2018} also tried 512-dimensional embeddings and reported no significant effects on the results.

\textbf{Multilabel Weight Imprinting (MWI):} The MWI layer is a single dense layer with an input size equal to the embedding size of the embeddings generator and output equal to the number of classes.
We initialize the MWI weights as an average of the embeddings for each class corresponding to the weight column.
We normalize the weights columns and use sigmoid activations with a threshold.
When training the MWI layer, we use the binary cross-entropy loss with an Adam optimizer \cite{kingma2014adam}. 

\textbf{MWI+ = MWI + Training (T) + Augmentations (A)}: When training with non-trivial (nt) augmentations, we use 3 types of augmentations \cite{chen_simple_2020}: i. random crop, resize, and random horizontal flip; ii. random color jitter; iii. random Gaussian blur. Trivial (t) augmentations refer to repeating the image. 

\textbf{Continual Learning (CL)} We use Experience Replay (ER) \cite{lin1992self}, which involves keeping a memory of old data and rehearsing it. ER has been used for CL \cite{rolnick2018experience, chaudhry2019tiny, hayes2019memory} and has been shown to outperform many CL approaches with and without a memory buffer \cite{chaudhry2019tiny}. In our multilabel continual learning setting, we retrain the old data with \textit{having/not having} the new label when new labels are received.

\section{Experiment Study}

\subsection{Datasets}
We selected 10 single-label and 5 multi-label datasets based on 5 reasons: i. \textit{Few-shot learning: } We added commonly used datasets for FSL; ii. \textit{Diversity:} We included diverse datasets to evaluate performance under distributional and task shifts; iii. \textit{Robustness:} We also picked an adversarial example dataset to evaluate robustness; iv. \textit{Multilabel settings:} We chose multilabel datasets, including object detection, fine-grained detection, and overlapping labels; v. \textit{AI for good:} We included a medical dataset to illustrate the broader impact of our work.
Our dataset list is in Table \ref{tab:datasets}.

\subsection{Evaluations} 

We compared FSL in 7 settings: i. using different embeddings generators; ii. using sigmoid (MWI) versus softmax (WI) activations; iii. with and without training (T) and augmentations (A), both trivial (t) and non-trivial (nt) augmentations; iv. in 4 FSL settings like \cite{sung_learning_2018}): (5-way 5-shot, 15 test; 20-way 5-shot, 5 test; 5 way 1 shot, 19 test; 20-way 1-shot, 10 test); v. in continual learning settings; vi. with CLIP's zero-shot and FSL linear probe; vii. with state-of-the-art (SOTA) results -- there are no previous few-shot, multi-label, and continual learning evaluations, but we compare with FSL and also full training/test set evaluations.
All evaluations are 5-way 5-shot, except for Ch (results in Appendix). We randomly sample classes and data points from each dataset 100 times and average the results.

\begin{table}[]
    \caption{Details about our chosen datasets, including their abbreviations (Abbr.).}
    \label{tab:datasets}
    \centering
    \begin{tabular}{lllll}
    Dataset Name & Abbr. & SOTA & Content & Selection Reason \\
    \hline
    \multicolumn{5}{l}{\textbf{Single}}
    \\
        Omniglot \cite{lake2015human} & OM & \cite{liu2019decoder} & Handwritten chars & Few-shot \\
        MiniImagenet \cite{vinyals_matching_2017} & MI & \cite{luo2019direction} &  ImageNet subset & Few-shot\\
        Labeled Faces in the Wild \cite{huang2008labeled} & LFW & \cite{chrysos2020p} & Faces & Diversity - Faces \\
        UCF101 \cite{soomro2012ucf101} & UCF & \cite{kalfaoglu2020late} & Action videos & Diversity - Actions \\
        Imagenet-R \cite{hendrycks2020many} & IR & \cite{radford2021learning} & ImageNet art & Diversity - Art \\
        Imagenet-Sketch \cite{wang2019learning} & IS & \cite{wang2019learning} & Imagenet sketches & Diversity - Sketch \\
        Indoor Scene Recognition \cite{quattoni2009recognizing} & ISR & \cite{rahimzadeh2021wise} & Indoor location & Diversity - Indoor \\
        CIFAR10 \cite{Krizhevsky09learningmultiple} & C10 & \cite{foret2020sharpness} & 10 classes & Diversity \\
        Imagenet-A \cite{hendrycks2019nae} & IA & \cite{radford2021learning} & Difficult images & Robustness \\
        Colorectal Histology \cite{kather2016multi} & CH & \cite{ghosh2021colorectal} &  Medical Images & AI for good \\
            \multicolumn{5}{l}{\textbf{Multi-label}}
    \\
        CelebA Attributes \cite{liu2018large} & CAA & \cite{kehrenberg2020null} & 40 attribute & Label overlap \\
        UTK Faces \cite{zhifei2017cvpr} & UTK & \cite{karkkainen2019fairface} & Gender, age, race & Label overlap\\
        Yale Faces \cite{georghiades2001few} & YF & \cite{khalili2019robust} & 11 face labels & Label overlap \\
        Common Objects in Context \cite{lin2014microsoft} & COCO & \cite{ridnik2021imagenet} & 90 objects & Object Detection \\
        iMaterialist (Fashion) \cite{guo2019imaterialist} & IM & \cite{guo2019imaterialist} &  Fashion/apparel & Fine-grained \\
        \hline
    \end{tabular}
\end{table}

\subsection{Metrics} 
Commonly used single-label classification metrics, e.g., top-1 accuracy, are not applicable in multilabel settings. Multilabel evaluations have used different metrics, including class and overall precision, recall, and F1, as well as mean average precision (mAP) \cite{wang2016cnn}. We calculated a total of 13 diverse metrics for each of our evaluations and included all the results in the supplementary materials.

We primarily use overall \textbf{F1-score} in this paper since F1-score accounts for class imbalance, which may be present in multilabel datasets, especially in real-world settings. The only downside of F1-score is that compared to mAP, it is threshold-dependent. However, in real-life situations, the threshold is also important, and therefore, we also discuss the optimal cut-off thresholds for our evaluations.

To compare our results with state-of-the-art (SOTA) results, we also report the metrics used by different SOTA results, i.e., top-1 accuracy for single-label datasets and average class accuracy (cAc) for multi-label datasets, except COCO, which uses mAP. We report these metrics along with F1-scores so that the F1-scores can be compared to the different SOTA metrics. SOTA references are in Table \ref{tab:datasets}.

\section{Results}

\begin{figure}
    \includegraphics[width=\linewidth]{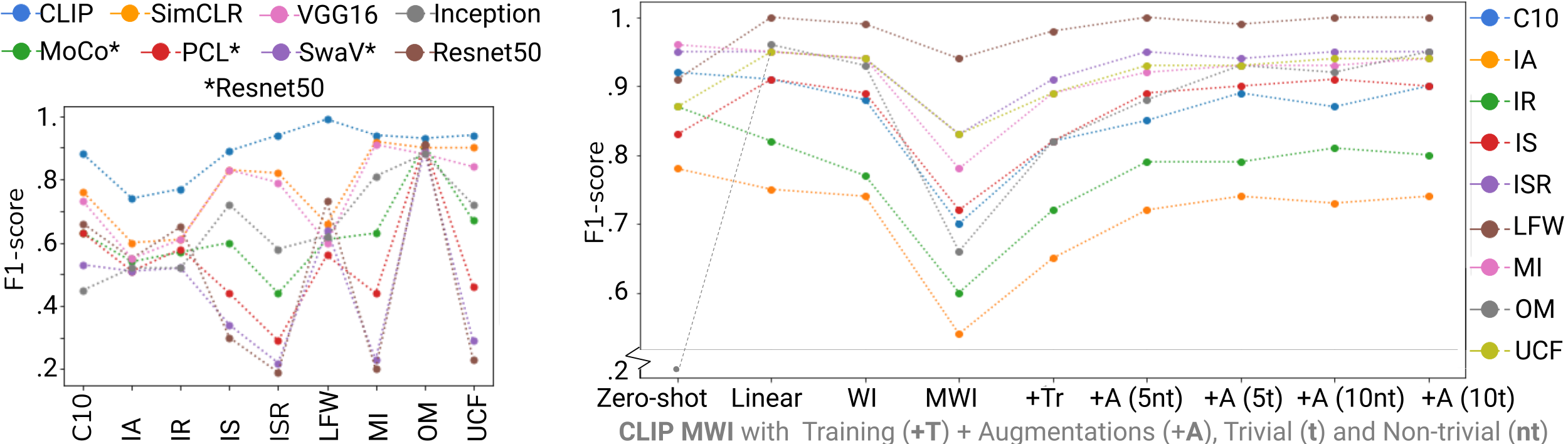}
    \caption{Comparing embeddings models (left) and Multilabel Weight Imprinting results (right).}
    \label{fig:fig1}
\end{figure}

\begin{figure}[h!]
    \includegraphics[width=\linewidth]{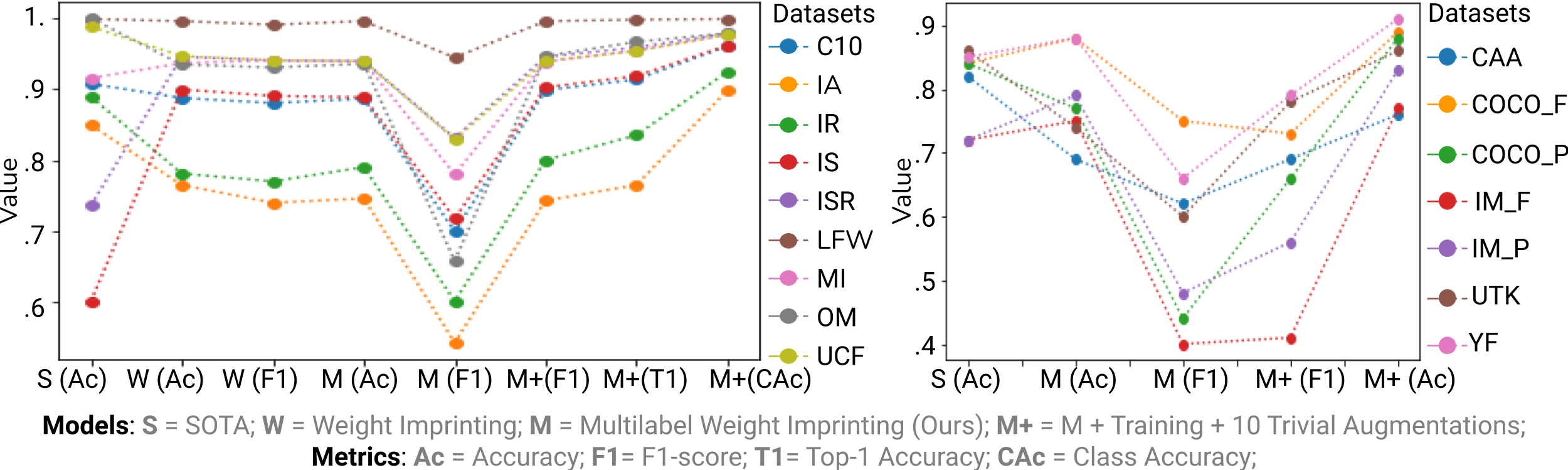}
    \caption{Comparing SOTA, WI, and MWI for single-label (left) and multi-label (right) datasets.}
    \label{fig:fig7}
\end{figure}

We share our main results in this section and ablations in the next. 
First, CLIP+WI performance is similar to CLIP's linear probe performance, possibly because both are linear layers (Fig \ref{fig:fig1}). Second, without training and augmentations, MWI performs worse than WI for single-label classifications -- MWI F1-score is worse than WI F1-score (Fig \ref{fig:fig1}), even though the accuracies are comparable (Fig \ref{fig:fig7}). Third, MWI with training (50-80 epochs) and augmentations (10 trivial/non-trivial), i.e., MWI+, does at least as well as WI using CLIP (Fig \ref{fig:fig1}-\ref{fig:fig7}), CLIP's linear probes (Fig -\ref{fig:fig1}), and state-of-the-art baselines (Fig \ref{fig:fig7}). MWI and MWI+ are also compared with SOTA and CLIP's baselines in Table \ref{tab:final}.

\begin{table}[]
\caption{Comparing CLIPPER Multilabel Weight Imprinting (MWI) with SOTA and CLIP baselines\textsuperscript{\ref{clip}}
}
\label{tab:final}
\centering
\begin{tabular}{llllllllllllllll}
\toprule
                  & C10  & Ia   & Ir   & Is   & Isr  & Lfw  & Mi & Om & Ucf  & Ca  & Co & Im & Ut & Yf & Ch \\
                             \hline
1        & .91 & .85 & .89 & .60 & .74 & 1.0  & .92 & 1.0  & .99 & .82 & .84 & .72 & .86 & .85 & .93 \\
2 & .91 & .75 & .82 & .91 & .95 & 1.0  & .95 & .96 & .95 \\
3         & .89 & .75 & .79 & .89 & .94 & 1.0  & .94 & .94 & .94 & .69 & .88 & .79 & .74 & .88\\
4          & .70 & .54 & .60 & .72 & .83 & .94 & .78 & .66 & .83 & .62 & .75 & .48 & .60 & .66 \\
5        & .90 & .74 & .80 & .90 & .94 & 1.0  & .94 & .95 & .94 & .69 & .73 & .56 & .78 & .79 & .69 \\
6 & .96  & .90   & .92   & .96   & .98  & .99  & .98   & .98   & .98  & .76  & .89 & .83   & .86  & .91 & .92 \\
7 & .91  & .77   & .83   & .92   & .96  & 1.0  & .95  & .97   & .95  & --  & .87 & --  & --  & -- \\

\multicolumn{16}{l}{\textbf{SOTA}(1); \textbf{CLIP} Lin. Probe Ac(2); \textbf{MWI} Ac(3),F1(4); \textbf{MWI+T+A} F1(5),CAc(6),Top1/mAP(7)}
\\
\bottomrule
\end{tabular}
\end{table}

\textbf{Comparing CLIP's embeddings: }
We compared embeddings from different pretrained supervised (Resnet50 \cite{he2016deep}, VGG16 \cite{simonyan2014very}, Inception V3), self-supervised (SimCLR v2 \cite{chen_big_2020}, MoCo v2 Resnet50 \cite{chen_simple_2020, chen2020improved}, PCL Resnet50 \cite{tang2018pcl}, SwAV Resnet50 \cite{caron2020unsupervised}), and weakly supervised (CLIP \cite{radford2021learning}) models (Figure \ref{fig:fig1} left). 
CLIP is trained on 400 million images, while the others are on 14 million Imagenet images.
We have three key findings.
First, CLIP gives the best results, possibly because CLIP is trained on a bigger dataset than other pretrained models.
Second, SimCLR's performance is closest to CLIP, even though it was trained only on a smaller dataset than CLIP. 
Third, Resnet50-based models performed much worse than the other models, even though all models, except CLIP, were trained on Imagenet.

\textbf{SOTA caveats:} There are 3 caveats to our SOTA comparisons (Table \ref{tab:final}). First, to the best of our knowledge, there is no prior work on multilabel few-shot learning and hence, we are setting new benchmarks and have no direct prior work to compare with. Second, even though we list the SOTA 5-way 5-shot results for OM and MI, there are two main differences: i. Previous few-shot results were pre-trained on large-scale, domain-specific, and labeled datasets, whereas our model is trained only on the few-shot set. Thus, performance for new domains like OM may not be as good as few-shot models pre-trained on OM; ii. Also, previous few-shot works did not do multilabel few-shot learning. Third, we list SOTA for other datasets, which have previously not been evaluated for few-shot learning, so the SOTA results are for full dataset training and testing and we only list them as a reference. \footnote{\label{clip}Both 1. Ia and Ir \& 2 in Table \ref{tab:final} use CLIP but 2 uses the ViT B/32 architecture whereas 1. Ia and Ir use the ViT L/14-336px architecture. L/14-336px is a bigger and better performing architecture but is not public \cite{radford2021learning}}

\textbf{CLIP baselines:} Since we use embeddings from CLIP, we also compare our CLIP + MWI results to CLIP's linear probe, zero-shot, and CLIP + WI performance. We use both F1-score and accuracy, and all comparisons, other than zero-shot, are 5-way 5-shot. MWI+ using CLIP is comparable to CLIP's baselines, but unlike the linear probe, also enables few-shot, multilabel, and continual learning.

\section{Ablations}

\subsection{MWI: Without Training and Augmentations}

\begin{figure}
    \includegraphics[width=\linewidth]{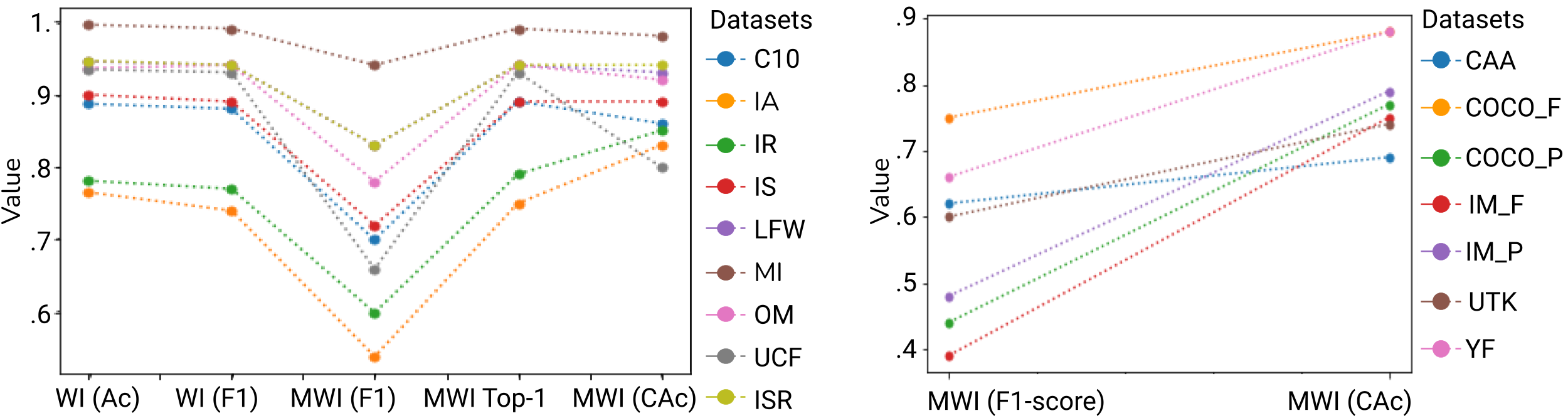}
    \caption{Comparing metrics, without training and augmentations, for single- and multi-label datasets.}
    \label{fig:fig2_3}
\end{figure}

We compare CLIPPER's 5-way 5-shot performance on 9 \textbf{single-label datasets} (Figure \ref{fig:fig2_3} left) and 5 \textbf{multilabel datasets} (Figure \ref{fig:fig2_3} right). For single-label, we perform two evaluations: i. Weight imprinting with softmax activations (f1-score and accuracy); ii. Multilabel weight imprinting with sigmoids (f1-score, top-1 accuracy, and per-class accuracy). For multilabel datasets with bounding boxes, i.e., COCO and iMaterialist, we compared full-full and patch-patch configurations, where `full' represents the full image and `patch' represents the bounding box of the relevant object. In n-m, n represents the training configuration and m represents the testing configuration. 

We had three key findings (Fig \ref{fig:fig2_3}). First, for single-label datasets, WI accuracy is comparable to MWI top-1 accuracy, which means that the sigmoid activation can get us comparable results to the softmax activation. Though, as expected, MWI F1-scores are much lower in value than MWI Top-1 accuracy. Second, multi-label datasets on average have much lower performance than single-label datasets, which is expected as they have more labels than single-label datasets. Third, the patch-patch configuration works best for iMaterialist whereas the full-full configuration works best for COCO, possibly because the background is meaningful in COCO but mostly white in iMaterialist. 

\subsection{MWI+: With Training and Augmentations}

\begin{figure}
    \includegraphics[width=\linewidth]{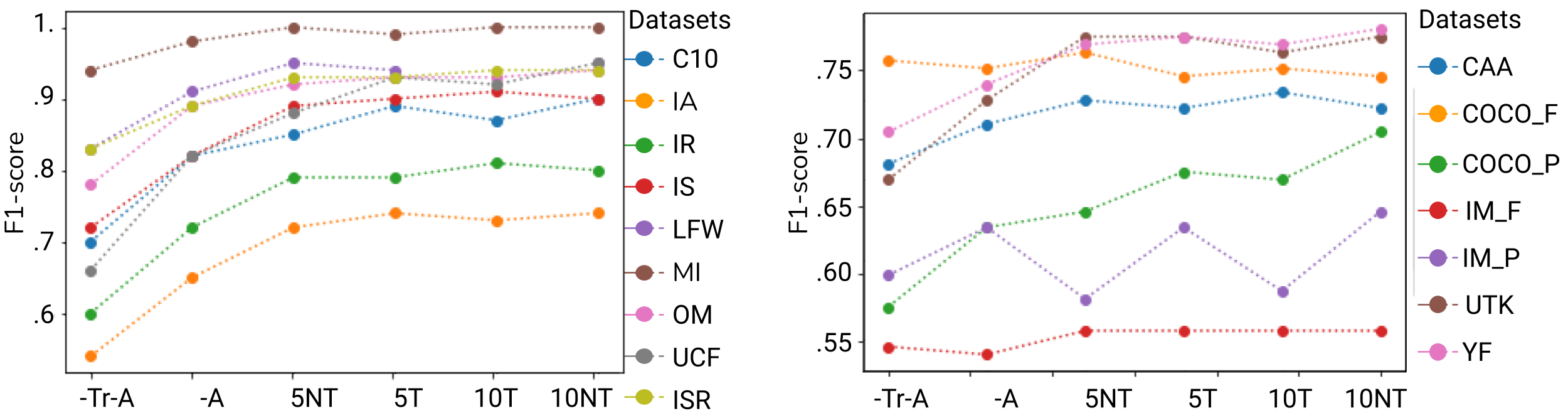}
    \caption{Comparing MWI with training and augmentations for single- and multi-label datasets}.
    \label{fig:fig4}
\end{figure}

\begin{figure}
    \includegraphics[width=\linewidth]{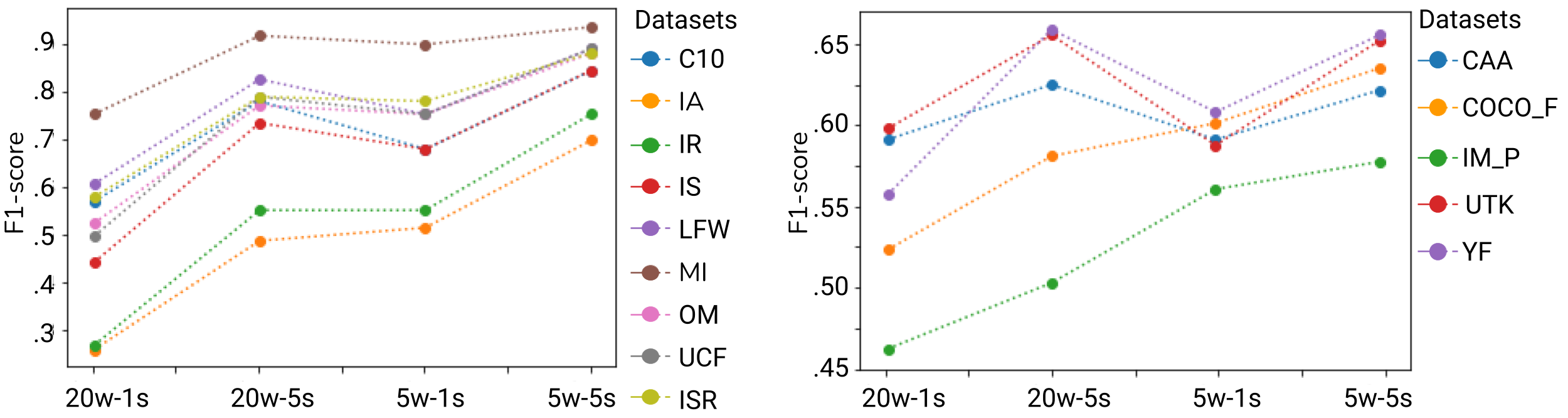}
    \caption{Comparing different few-shot settings with MWI+ (L:single-label, R:multi-label datasets)}
    \label{fig:fig5}
\end{figure}

We evaluated the performance of Multilabel Weight Imprinting by adding \textbf{training (T) and augmentations (A)} (Figure \ref{fig:fig4}). We had three key findings. First, CLIPPER's performance improved with both training and augmentations -- after training and augmentations, F1-scores for multi-label weight imprinting were comparable to the F1-scores for weight imprinting with softmax. Second, the performance saturates around 50-80 epochs, and trivial (t) augmentations, i.e., image repetitions, are as good or sometimes even better than non-trivial (nt) augmentations. Third, with training, the best threshold values stabilized around 0.5 for most datasets (Figure \ref{fig:thresh} (left). We also compared 4 \textbf{few-shot learning settings} (Figure \ref{fig:fig5}): 5-way 5-shot, 20-way 5-shot, 5 way 1 shot, 20-way 1-shot. The performance worsens with decreasing shots and with increasing classes.

\begin{figure}
    \includegraphics[width=\linewidth]{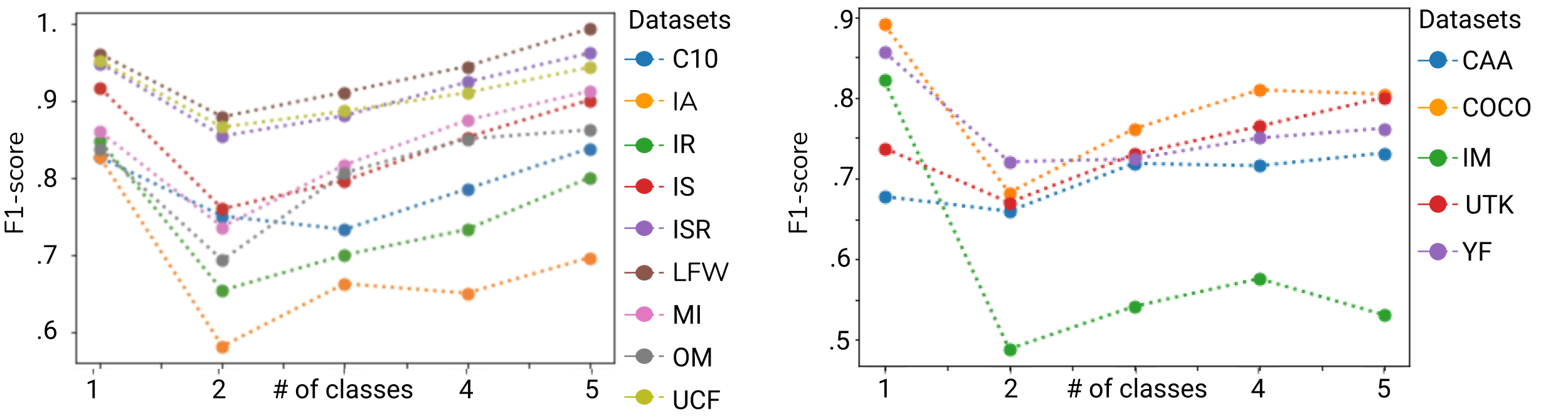}
    \caption{MWI+ Continual learning results for increasing classes (L:single-, R:multi-label dataset)}
    \label{fig:fig6}
\end{figure}

\begin{figure}[h!]
    \includegraphics[width=\linewidth]{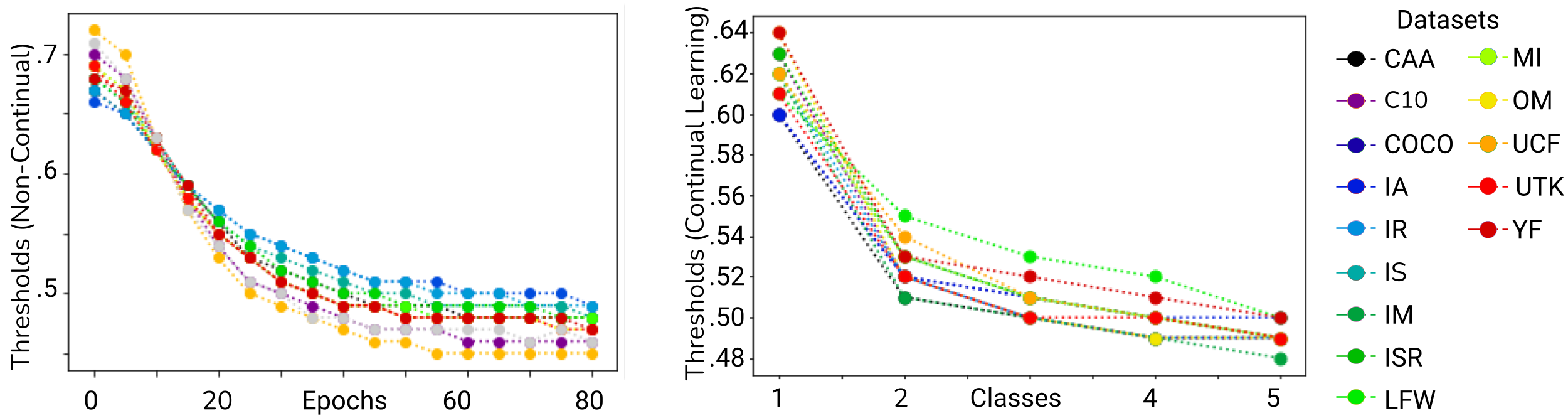}
    \caption{Optimal thresholds with (left) and without (right) continual learning for all datasets.}
    \label{fig:thresh}
\end{figure}

\subsection{Continual learning}
We evaluated 5 way 5 shot continual learning. We incrementally added the number of labeled classes and their respective training data and labels, while keeping the test set fixed. We had three key findings. First, CL performance (Figure \ref{fig:fig6}) varies with the number of classes but reaches approximately the same 5-way 5-shot value with continual learning as it does without continual learning (\ref{fig:fig4}). Second, the optimal-performance thresholds vary with the number of classes and we share the best accuracies and their respective thresholds for each dataset for different number of classes (Figure \ref{fig:thresh} right). Third, the thresholds are higher with lower number of classes, possibly because of lesser training data, but converge to approximately the same 5-way 5-shot value with and without continual learning (Fig \ref{fig:thresh}).

\section{Discussion and Limitations} 

With advances in representation learning, either using generative or contrastive models, the question arises: \textit{how to best use the representations in downstream tasks}. Previous work suggests, ``combining the strength of zero-shot transfer with the flexibility of few-shot learning is a promising direction'' \cite{radford2021learning} and ``obtain better results...by combining few-shot learning methods with fine-tuning'' \cite{kornblith2019better}.

We outline few-shot, continual, and multilabel learning as the desiderata for downstream tasks and introduce a technique, called Multilabel Weight Imprinting, to meet the desiderata. Our model uses embeddings from a pertained CLIP model and shows promising performance on diverse and challenging tasks. We set few-shot, multilabel, and continual learning benchmarks for many datasets.

Our work has three \textbf{key findings}. First, using pretrained models with an existing FSL technique, i.e., weight imprinting \cite{qi_low-shot_2018}, enables sample-efficient learning with two additional benefits: i. Unlike commonly-used transfer learning techniques like fine-tuning and distillation, we have a prototype for each class and can flexibly add/update each class prototype without influencing (e.g., forgetting) the other class prototypes; ii. Unlike commonly-used FSL methods, the base model need not be trained with computationally-intensive techniques involving large, domain-specific, and expensive-to-label datasets. Second, replacing weight imprinting's softmax function with a sigmoid and threshold function enables multilabel weight imprinting, and using training and augmentations helps improve performance. Finally, adding experience reply enables multilabel continual learning.

Our work has three \textbf{key limitations}: i. Multilabel learning has poorer and even threshold-dependent performance compared to single-label learning, but multi-label learning is still more realistic than single-label classification as even single-label datasets have multiple labels \cite{yun2021re}; ii. Prototype-based few-shot learning scales the number of prototypes with the number of classes and comparing with every single prototype may not be efficient. Thus, efficient and scalable methods, e.g., hierarchical prototypes, are needed; iii. Experience replay for multilabel continual learning is memory-inefficient and memory-efficient continual learning, e.g., prototype-based contrastive learning, could be leveraged.

We have three key directions for \textbf{future work}: i. Use downstream few-shot learning for correcting/filtering labels from upstream models, without necessarily sending data to the upstream models; ii. Make few-shot, multilabel, and continual learning memory-efficient, robust, and deployable; iii. Deploy and test in real-world settings, e.g., for human-in-the-loop and personalized applications.

\section{Broader Impact}
We highlight three key areas for positive impact of our work. First, we designed our model for few-shot, multilabel, and continual learning to enable real-world sample-efficient applications, including personalized and AI for good applications (more details in appendix). Second, since we do not train the upstream model, the data does not have to be sent to the upstream model, affording privacy-preserving and offline model training. Third, since we only train a linear layer, our model affords easy and lightweight real-world training and deployment, including on mobile, web, and wearable devices, especially if the pretrained base model can be mobile-optimized \cite{howard2017mobilenets, tan2019efficientnet} as in \cite{khan2021cvpr}.

Like many machine learning techniques, our model is as good as its applications. 
We have made our model flexible, easy-to-use, and easy-to-train -- it can be used with any state-of-the-art pre-trained model, trained and run using free Google Colab notebooks, and personalized using only a few examples.
Moreover, few-shot and personalized learning may also help mitigate data/labeling bias.

Thus, we hope that our work will enable interdisciplinary stakeholders to ethically design and deploy personalized, privacy-preserving, and meaningful real-world deep learning applications.

\section{Conclusion}

Data-efficiency and generalization are key challenges for deep learning. Self-supervised or weakly supervised models trained on unlabeled or uncurated datasets have shown promising transfer to few-shot tasks. Few-shot learning methods have also demonstrated sample-efficient learning.

We highlight the need for few-shot, multilabel, and continual learning, and developed a technique, called Multi-label Weight Imprinting (MWI), for few-shot, continual, and multi-label learning.
Unlike previous FSL techniques, our model, CLIPPER, uses MWI with pretrained representations from a weakly-supervised model, i.e., CLIP. Thus, CLIPPER combines the sample-efficiency and generalizability of transfer learning with the flexibility and specialization of FSL methods.

CLIPPER shows robust and competitive performance and is the first step in the direction of using pretrained models for few-shot, multilabel, and continual learning.
Our model is also lightweight and the data does not have to be sent back to the upstream model, enabling privacy-preserving and on-device downstream training. Thus, our model enables few-shot, multilabel, and continual learning, especially for easy-to-train, light-weight, and privacy-preserving real-world applications.

\bibliography{main}
\bibliographystyle{abbrv}

\pagebreak

\appendix

\section{Method and Algorithm}\label{algorithm}

We describe our problem definition below and the algorithm in Algorithms \ref{augmentations} - \ref{continual_learning}.


We have a train set of N labeled examples: $S =\{(x_1, y_1), . . . , (x_N , y_N )\}$ where each $x_i \in \mathbb{R}^D$  is the D-dimensional feature vector of an example and $y_i$ is a subset of labels $L = \{1, . . . , K\}$. $S_k$ denotes the set of examples labeled with class k and $S_j$ and $S_k$ are not necessarily disjoint sets for $ j \neq k$. 

Moreover, in continual learning setting, the training set $S$ and the labels $L$ can grow over time. $S^t =\{(x_1, l_1), . . . , (x_M , l_M )\}$ and $L^t = \{1, . . . , P\}$ and  $S^t =\{(x_1, l_1), . . . , (x_N , l_N )\}$ and $L^t = \{1, . . . , Q\}$ where $N > M$ and $Q > P$, i.e., the size of labels and the training set are non-decreasing over time.
 
\begin{algorithm}

\caption{Augmentations Function}
\label{augmentations}

\begin{tabularx}{\textwidth}{l<{}X}
\hspace*{\algorithmicindent} \textbf{Input}: & Original image, \textbf{x} \\
\hspace*{\algorithmicindent} \textbf{Output}: & Augmented image, \textbf{x'}
\end{tabularx}

\begin{algorithmic}[1]

\Function{Augment}{$\mathbf{x}$}
\State $\mathbf{x'} = \Call{RandomCropAndResize}{\mathbf{x}}$
\State $\mathbf{x'} = \Call{RandomHorizontalFlip}{\mathbf{x'}}$
\State $\mathbf{x'} = \Call{RandomColorJitter}{\mathbf{x'}}$
\State $\mathbf{x'} = \Call{RandomGaussianBlur}{\mathbf{x'}}$

\State \Return $\mathbf{x'}$
\EndFunction

\end{algorithmic}
\end{algorithm}

\begin{algorithm}
\caption{Embedding Function}
\label{embedding}

\begin{tabularx}{\textwidth}{l<{}X}
\hspace*{\algorithmicindent} \textbf{Input}: & Dataset to embed $D$, Number of augmentations for each image $N_A$ \\
\hspace*{\algorithmicindent} \textbf{Output}: & Embedded dataset $E$
\end{tabularx}

\begin{algorithmic}[1]

\Function{CLIPEmbedDataset}{$D, N_A$}
\State $E \gets \{\}$ \Comment{Embedded dataset}
\For{$(\mathbf{x_i}, y_i) \in D$}
\For{$j \in \{1, 2, \cdots N_A\}$}
\State $\mathbf{x_i'} \gets \Call{Augment}{\mathbf{x_i}}$
\Comment{If augmenting(ablation)}
\State $\phi_i \gets \Call{Clip}{\mathbf{\mathbf{x_i'}}}$
\State $\phi_i' \gets \Call{Normalize}{\phi_i}$
\State $\Call{Append}{E, (\phi_i', y_i)}$
\EndFor
\EndFor
\State \Return{$E$}
\EndFunction

\end{algorithmic}
\end{algorithm}

\begin{algorithm}
\caption{Training Function}
\label{training}

\begin{tabularx}{\textwidth}{l<{}X}
\hspace*{\algorithmicindent} \textbf{Input}: & Embedded dataset $E$, Weight Imprinting single layer $f$, Number of training epochs $e_{train}$\\
\hspace*{\algorithmicindent} \textbf{Output}: & Trained weight imprinting layer $f$
\end{tabularx}

\begin{algorithmic}[1]

\Function{Train}{$E, f, e_{train}$}
\State $J \gets 0$
\For{$(\phi_i', y_i) \in E$}
\State $y_i' = \sigma(f(\phi_i')$
\State $J \gets J + \Call{BinaryCrossentropy}{y_i, y_i'}$
\EndFor
\State $J \gets \Call{Mean}{J}$
\State $\Call{Optimize}{f, J, Adam, e_{train}}$
\State \Return $f$
\EndFunction

\end{algorithmic}
\end{algorithm}

\begin{algorithm}

\caption{Predicting Function}
\label{prediction}

\begin{tabularx}{\textwidth}{l<{}X}
\hspace*{\algorithmicindent} \textbf{Input}: & Embedding $\phi'$,  Weight Imprinting single layer $f$, Evaluation threshold $threshold$\\
\hspace*{\algorithmicindent} \textbf{Output}: & Predicted labels $labels$
\end{tabularx}

\begin{algorithmic}[1]

\Function{Predict}{$\phi', f, threshold$}
\State $labels \gets \{\}$
\State $y_i' = \sigma(f(\phi_i')$
\For{$j \in \{1, 2, \cdots W\}$}
\If{$y_{ij}' \geq threshold$}
\State \Call{Append}{$labels, j$}
\EndIf
\EndFor
\State \Return $labels$
\EndFunction

\end{algorithmic}
\end{algorithm}

\begin{algorithm}

\begin{tabularx}{\textwidth}{l<{}X}
\hspace*{\algorithmicindent} \textbf{Input}: & Number of ways/labels $n_{ways}$, Number of shots per label $n_{shot}$, Number of test images per label $n_{test}$, Number of episodes $n_{episodes}$, Dataset to sample from $D$\\
\hspace*{\algorithmicindent} \textbf{Output}: & Few shot train dataset $D'_{train}$, Few shot test dataset $D'_{test}$
\end{tabularx}

\begin{algorithmic}[1]
\caption{Sampling Function}
\label{sampling}

\Function{Sample}{$n_{way}, n_{shot}, n_{test}, n_{episodes}, D$}
\State $D'_{train} = \{\}$
\State $D'_{test} = \{\}$
\State $L \gets \Call{GetUniqueLabels}{D}$
\For{$i \in \{1, 2, \cdots n_{episodes}\}$}
\State $d'_{train} \gets \{\}$
\State $d'_{test} \gets \{\}$
\State $L_{sampled} \gets \Call{RandomSample}{L, n_{way}}$
\Comment{Sample $n_{way}$ labels}
\For{$l \in L_{sampled}$}
\State $D_l \gets \{(\mathbf{x_i}, y_i) \in D \mid l \in y_i \}$
\State $d_1 \gets \Call{RandomSample}{D_l \setminus ( d'_{train} \cup d'_{test}), n_{shot}}$
\State $\Call{Append}{d'_{train}, d_1}$
\State $d_2 \gets \Call{RandomSample}{D_l \setminus ( d'_{train} \cup d'_{test} \cup d_1), n_{test}}$
\State $\Call{Append}{d'_{test}, d_2}$
\EndFor
\State $\Call{Append}{D'_{test}, d'_{test}}$
\State $\Call{Append}{D'_{train}, d'_{train}}$
\EndFor
\State \Return $D'_{train}, D'_{test}$
\EndFunction

\end{algorithmic}
\end{algorithm}

\begin{algorithm}

\begin{tabularx}{\textwidth}{l<{}X}
\hspace*{\algorithmicindent} \textbf{Input}: & Embedded few shot train dataset $E_{train}$, Embedded few shot test dataset $E_{test}$, Evaluation threshold $threshold$, Number of training epochs $e_{train}$
\end{tabularx}

\begin{algorithmic}[1]

\caption{Continual Learning Evaluation}
\label{continual_learning}

\Function{ContinualLearning}{$E_{train}, E_{test}, threshold, e_{train}$}
\State $L \gets \Call{GetUniqueLabels}{E_{train}}$
\State $LabelsAdded = \{\}$
\For{$l \in L$}
\State $E_l \gets \{(\phi_i', l) \mid (\phi_i', y_i) \in E_{train}, l \in y_i\}$
\If{\Call{IsEmpty}{$LabelsAdded$}}
\State $f = \Call{InitializeWeightImprinting}{E_l}$
\EndIf
\If{!\Call{IsEmpty}{$LabelsAdded$}}
\State $f = \Call{AddClassWeightImprinting}{f, E_l}$
\EndIf

\State $E'_{train} \gets \{ (\phi_i', \{l_j\}) \mid (\phi_i', y_i) \in E_{train}, l_j \in LabelsAdded \}$
\State \Call{Train}{$E'_{train}, f, e_{train}$}

\State $E'_{test} \gets \{ (\phi_i', \{l_j\}) \mid (\phi_i', y_i) \in E_{test}, l_j \in LabelsAdded \}$
\For{$(\phi_i', y_i) \in E'_{test}$}
\State $labels \gets \Call{Predict}{\phi_i', f, threshold}$
\State \Call{EvaluateMetric}{$labels, y_i$}
\EndFor
\State \Call{Append}{$LabelsAdded, l$}

\EndFor
\EndFunction

\end{algorithmic}
\end{algorithm}

\pagebreak

\section{Real-world Applications}\label{applications}
We discuss the real-world applications of our work using a personalized photo-labeling application and few-shot results for a medical imaging dataset.

\begin{figure}
    \centering
    \includegraphics[width=\linewidth]{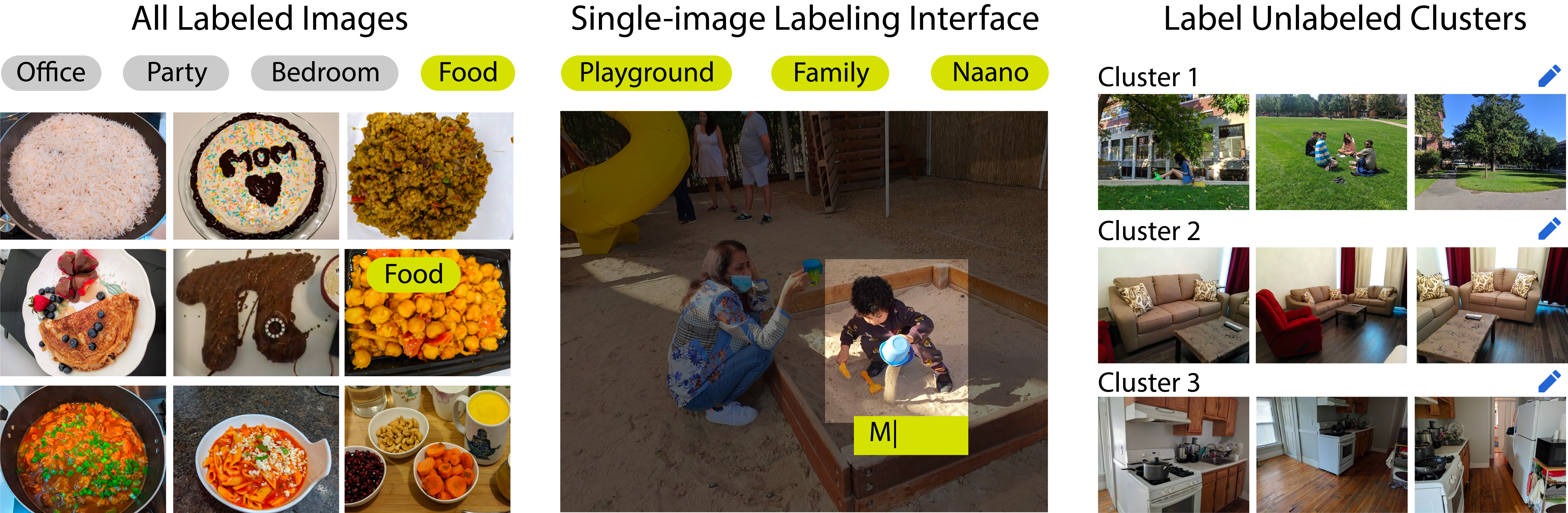}
    \caption{Design for personalized photo labeling application, showing all labeled images, image-labeling interface, and cluster labeling interface}
    \label{fig:application}
\end{figure}

\begin{figure}
    \centering
    \includegraphics[width=\linewidth]{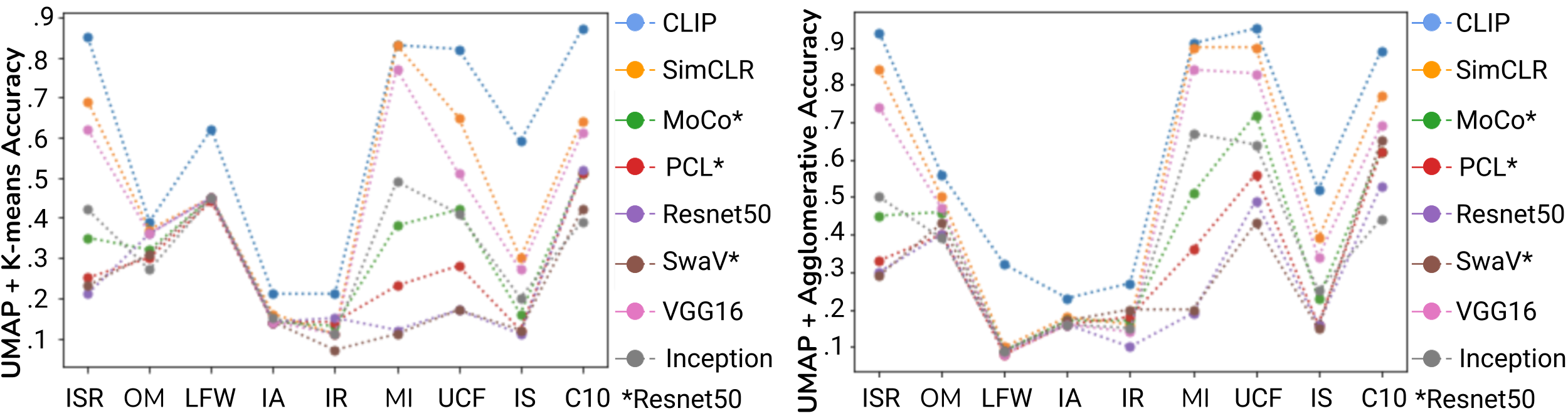}
    \caption{Clustering using different embeddings: Kmeans (Left), Aggolomerate Clustering (Right)}
    \label{fig:clustering}
\end{figure}

\textbf{Personalized photo labeling:}
We designed a personalized photo labeling application to enable personalized phot labeling using CLIPPER's. The interface (Figure \ref{fig:application}) shows how users can label specific images and also, how clustering can help efficient labeling by clustering similar images.
We share the clustering accuracies in Figure \ref{fig:clustering} --
image embeddings from each embedding generator were passed to UMAP \cite{mcinnes2018umap}, resulting in 2-dimensional embeddings, which we used for clustering.

\begin{figure}
    \centering
    \includegraphics[width=\linewidth]{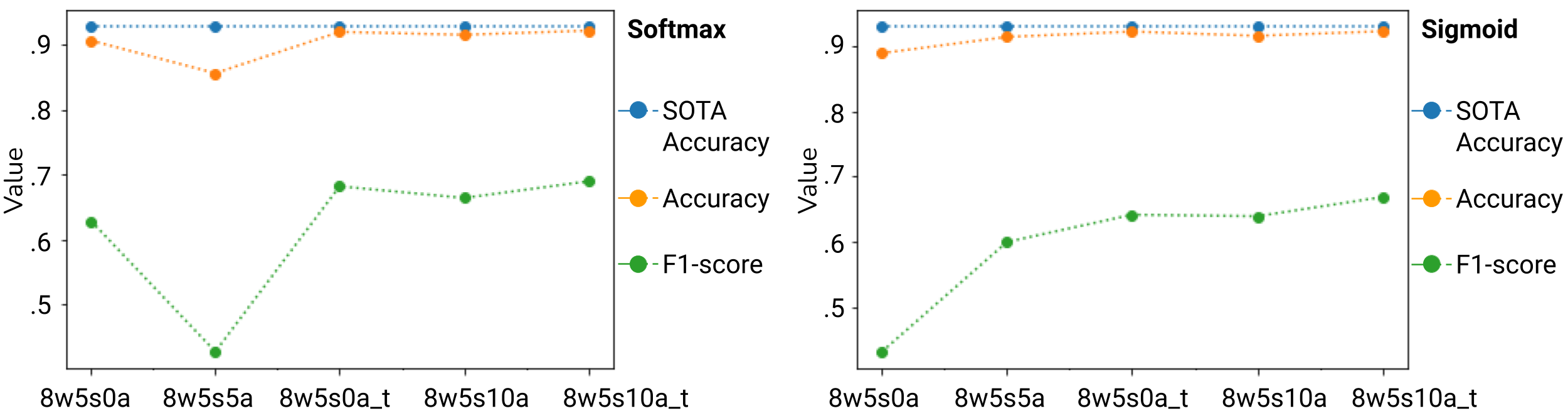}
    \caption{Comparing CLIPPER's FSL for colorectal cancer histology with state-of-the-art}
    \label{fig:ai_for_good}
\end{figure}

\textbf{Real-world Medical Imaging:}
We tested multilabel few-shot learning on the colorectal cancer histology dataset \cite{kather2016multi}. Our model, trained with 5 examples per class, shows competitive performance compared to the state-of-the-art model, which was trained using the full dataset \cite{ghosh2021colorectal} (Fig \ref{fig:ai_for_good}). 

\pagebreak

\section{Experiments and Results}

We share the copyright details for our datasets in Table \ref{tab:datasets_cap} and our metric details in Table \ref{tab:metrics_tab}.

For UCF and ISR, the images are resized such that the smaller dimension is 224 followed by a center crop of 224x224. For all other datasets, the images are resized to 224x224. Since UCF has videos, we select the middle frame as the target image. 

In the rest of the tables, we show the detailed numerical results for all our evaluations, including error bars (Table \ref{tab:error_bars}) for our few-shot evaluations.

\begin{table}[]
    \caption{Copyright details about our chosen datasets.}
    \label{tab:datasets_cap}
    \centering
    \begin{tabular}{ll}
    \toprule
    Dataset & Copyright \\
    \bottomrule
    \multicolumn{2}{l}{\textbf{Single}}
    \\
        Omniglot \cite{lake2015human} & MIT License Copyright (c) 2015 Brenden Lake \\
        MiniImagenet \cite{vinyals_matching_2017} & MIT License Copyright (c) 2019 Yaoyao Liu \\
        Labeled Faces in the Wild \cite{huang2008labeled} & Unknown \\
        UCF101 \cite{soomro2012ucf101} & Unknown \\
        Imagenet-R \cite{hendrycks2020many} & MIT License Copyright (c) 2020 Dan Hendrycks \\
        Imagenet-Sketch \cite{wang2019learning} & MIT License Copyright (c) 2021 Haohan Wang\\
        Indoor Scene Recognition \cite{quattoni2009recognizing} & Uknown \\
        CIFAR10 \cite{Krizhevsky09learningmultiple} & MIT License Copyright (c) 2017 Baptiste Wicht\\
        Imagenet-A \cite{hendrycks2019nae} & MIT License Copyright (c) 2019 Dan Hendrycks\\
        Colorectal Histology \cite{kather2016multi} & Creative Commons Attribution 4.0 International \\
            \multicolumn{2}{l}{\textbf{Multi-label}}
    \\
        CelebA Attributes \cite{liu2018large} & ``non-commercial research purposes''\\
        UTK Faces \cite{zhifei2017cvpr} & ``non-commercial research purposes''\\
        Yale Faces \cite{yaleface} & Data files © Original Authors \\
        Common Objects in Context \cite{lin2014microsoft} & Creative Commons Attribution 4.0 License. \\
        iMaterialist (Fashion) \cite{guo2019imaterialist} & MIT License Copyright (c) 2017 Visipedia\\
        \bottomrule
    \end{tabular}
\end{table}

\begin{table}[]
    \caption{Details of metrics used for single-label and multi-label classifications.}
    \label{tab:metrics_tab}
    \centering
    \begin{tabular}{lccc}
    \toprule
    Metrics & Single-label Softmax & Single-label Sigmoid & Multi-label \\
    \bottomrule
         Hamming score &  & \checkmark & \checkmark \\ 
         Jaccard & & \checkmark & \checkmark\\
         Subset accuracy & & \checkmark & \checkmark \\
         Mean Average Accuracy (mAP) & & \checkmark & \checkmark\\
         Class F1 & \checkmark & \checkmark & \checkmark \\
         Overall F1 & \checkmark & \checkmark & \checkmark \\
         Class precision & \checkmark & \checkmark & \checkmark \\
         Overall precision & \checkmark & \checkmark & \checkmark \\
         Class recall & \checkmark & \checkmark & \checkmark \\
         Overall recall & \checkmark & \checkmark & \checkmark \\
         Top-1 accuracy & \checkmark & \checkmark &  \\
         Top-5 accuracy & \checkmark & \checkmark &  \\
         Class accuracy & \checkmark & \checkmark & \checkmark \\
         \bottomrule
    \end{tabular}
    \label{tab:metrics}
\end{table}

\makeatletter
\csvset{
  autotabularcenter/.style={
    file=#1,
    after head=\csv@pretable\begin{tabular}{|*{\csv@columncount}{c|}}\csv@tablehead,
    table head=\hline\csvlinetotablerow\\\hline,
    late after line=\\,
    table foot=\\\hline,
    late after last line=\csv@tablefoot\end{tabular}\csv@posttable,
    command=\csvlinetotablerow},
  autobooktabularcenter/.style={
    file=#1,
    after head=\csv@pretable\begin{tabular}{*{\csv@columncount}{l}}\csv@tablehead,
    table head=\toprule\csvlinetotablerow\\\midrule,
    late after line=\\,
    table foot=\\\bottomrule,
    late after last line=\csv@tablefoot\end{tabular}\csv@posttable,
    command=\csvlinetotablerow},
}
\makeatother
\newcommand{\csvautotabularcenter}[2][]{\csvloop{autotabularcenter={#2},#1}}
\newcommand{\csvautobooktabularcenter}[2][]{\csvloop{autobooktabularcenter={#2},#1}}

\begin{table}
\centering
\caption{Comparing FSL (F1-score) using Weight Imprinting for different embeddings models}
\csvautobooktabularcenter[]{Final_CSVs/Figure-2-1.csv}
\vspace{1em}
\end{table}

\begin{table}
\centering
\caption{Comparing Multilabel Weight Imprinting F1-scores for single-label datasets}
\csvautobooktabularcenter[]{Final_CSVs/Figure-2-2.csv}
\vspace{1em}
\end{table}

\begin{table}
\centering
\caption{Comparing Multilabel Weight Imprinting, SOTA, and WI for single-label datasets}
\csvautobooktabularcenter[]{Final_CSVs/Figure-3-1.csv}
\vspace{1em}
\end{table}

\begin{table}
\centering
\caption{Comparing Multilabel Weight Imprinting, SOTA, and WI for multi-label datasets}
\csvautobooktabularcenter[respect underscore=true]{Final_CSVs/Figure-3-2.csv}
\end{table}

\begin{table}
\centering
\caption{Comparing metrics, without training and augmentations, for single-label datasets}
\csvautobooktabularcenter[]{Final_CSVs/Figure-4-1.csv}
\end{table}

\begin{table}
\centering
\caption{Comparing metrics, without training and augmentations, for multi-label datasets}
\csvautobooktabularcenter[respect underscore=true]{Final_CSVs/Figure-4-2.csv}
\end{table}

\begin{table}
\centering
\caption{Comparing MWI with training and augmentations for single-label datasets}
\csvautobooktabularcenter[]{Final_CSVs/Figure-5-1.csv}
\end{table}

\begin{table}
\centering
\caption{Comparing MWI with training and augmentations for multi-label datasets}
\csvautobooktabularcenter[respect underscore=true]{Final_CSVs/Figure-5-2.csv}
\end{table}

\begin{table}[]
\centering
\caption{Comparing CLIPPER's FSL for colorectal cancer histology with state-of-the-art}
\begin{tabular}{llllll}
\toprule
                 &      & \multicolumn{2}{l}{Single-label Evaluation} & \multicolumn{2}{l}{Multi-label Evaluation} \\
                 & SOTA & Accuracy            & Overall F1            & Accuracy            & Overall F1           \\
                 \bottomrule
8w5s0a           & 0.93 & 0.90                & 0.63                  & 0.89                & 0.43                 \\
8w5s5a           & 0.93 & 0.85                & 0.43                  & 0.91                & 0.6                  \\
8w5s5a\_trivial  & 0.93 & 0.92                & 0.68                  & 0.92                & 0.64                 \\
8w5s10a          & 0.93 & 0.92                & 0.66                  & 0.91                & 0.64                 \\
8w5s10a\_trivial & 0.93 & 0.92                & 0.69                  & 0.92                & 0.67                \\
\bottomrule
\end{tabular}
\end{table}

\begin{table}
\centering
\label{tab:error_bars}
\caption{Best performance (10 trivial augmentations) on all datasets for 5-way 5-shot with training - mean metric values with error across 100 episodes.}
\csvautobooktabularcenter[]{Final_CSVs/ErrorBars.csv}
\end{table}

\begin{table}
\centering
\caption{Comparing different few-shot settings with MWI+ for single-label datasets}
\csvautobooktabularcenter[]{Final_CSVs/Figure-6-1.csv}
\end{table}

\begin{table}
\centering
\caption{Comparing different few-shot settings with MWI+ for multi-label datasets}
\csvautobooktabularcenter[respect underscore=true]{Final_CSVs/Figure-6-2.csv}
\end{table}

\begin{table}
\centering
\caption{MWI+ Continual learning results for increasing classes for single-label datasets}
\csvautobooktabularcenter[respect underscore=true]{Final_CSVs/Figure-7-1.csv}
\end{table}

\begin{table}
\centering
\caption{MWI+ Continual learning results for increasing classes for multi-label datasets}
\csvautobooktabularcenter[respect underscore=true]{Final_CSVs/Figure-7-2.csv}
\end{table}

\begin{table}
\centering
\caption{Optimal thresholds without continual learning for single-label datasets}
\csvautobooktabularcenter[]{Final_CSVs/Figure-8-1-single.csv}
\end{table}

\begin{table}
\centering
\caption{Optimal thresholds without continual learning for multi-label datasets}
\csvautobooktabularcenter[respect underscore=true]{Final_CSVs/Figure-8-1-multi.csv}
\end{table}


\begin{table}[]
\centering
\caption{Optimal thresholds with continual learning for single- and multi-label datasets}
\begin{tabular}{lllllllllllllll}
\toprule
  & ca  & c10  & co & ia   & ir   & is   & im   & isr  & lfw  & mi   & om   & ucf  & utk  & yf   \\
\bottomrule
1 & .60  & .67 & .60  & .64 & .66 & .66 & .62 & .67 & .65 & .66 & .64 & .66 & .61 & .64 \\
2 & .51 & .56 & .52 & .56 & .56 & .57 & .51 & .57 & .59 & .57 & .52 & .58 & .52 & .53 \\
3 & .50  & .54 & .51 & .55 & .54 & .55 & .5  & .55 & .57 & .55 & .50  & .55 & .50  & .52 \\
4 & .49 & .53 & .50  & .54 & .53 & .54 & .49 & .54 & .56 & .54 & .49 & .54 & .50  & .51 \\
5 & .49 & .53 & .49 & .54 & .53 & .53 & .48 & .53 & .54 & .53 & .49 & .53 & .49 & .50 \\
\bottomrule
\end{tabular}
\end{table}

\end{document}